\documentclass[letterpaper, 10 pt, conference]{ieeeconf}

\IEEEoverridecommandlockouts
\usepackage{cite}
\usepackage{amsmath,amssymb,amsfonts}
\usepackage{algorithmic}
\usepackage{graphicx}
\usepackage{textcomp}
\usepackage{xcolor}
\usepackage{hyperref}
\hypersetup{hidelinks}
\usepackage{booktabs} 		% For standard tabular tables, with rules
\usepackage{tabularx}	    % For clean tables such as in the Nomenclature
\usepackage[capitalise]{cleveref}
\usepackage{subcaption}     % To arrange captions of subfigures
\usepackage{multirow}
\usepackage{array}
\usepackage{booktabs}
\usepackage[per-mode=symbol,exponent-product=\cdot]{siunitx}
\usepackage{float}			% Arranging of figures on the page
\usepackage{tikz}
\usepackage{graphicx}
\usepackage{pgfplots}
\usepackage{flushend}
\pgfplotsset{width=9cm,compat=1.8}

\def\BibTeX{{\rm B\kern-.05em{\sc i\kern-.025em b}\kern-.08em
    T\kern-.1667em\lower.7ex\hbox{E}\kern-.125emX}}

\title{\LARGE \bf Scenario Extraction from a Large Real-World Dataset for the Assessment of Automated Vehicles}

\author{Detian Guo$^{a,b}$, Manuel Muñoz Sánchez$^a*$, Erwin de Gelder$^{b}$, Tom P.J.\ van der Sande$^a$% <-this % stops a space
\thanks{This work was partially supported by SAFE-UP under EU's Horizon 2020 research and innovation programme, grant agreement 861570.}% <-this % stops a space
\thanks{$^{a}$Department of Mechanical Engineering, Eindhoven University of Technology, The Netherlands.}%
\thanks{$^{b}$Integrated Vehicle Safety, TNO, Helmond, The Netherlands.}
\thanks{$^*$Corresponding author:\ {\tt\small m.munoz.sanchez@tue.nl}}
}
\newcommand{\cstart}{\color{black}}
\newcommand{\cend}{\color{black}}

\newcommand{\mcB}{\mathcal{B}}

\newcommand{\mcBe}{\mcB_{\mathrm{e}}}
\newcommand{\mcBp}{\mcB_{\mathrm{p}}}
\newcommand{\currenttimestep}{k_{\mathrm{c}}}
\newcommand{\extendinghorizon}{T_{\mathrm{e}}}
\newcommand{\predictionhorizon}{T_{\mathrm{p}}}
\newcommand{\timeintervalturning}{T_{\mathrm{d}}}
\newcommand{\timestep}{T_{\mathrm{s}}}
\newcommand{\timestepend}{k_{\mathrm{e}}}
\newcommand{\timesteppredict}{k_{\mathrm{p}}}

\begin{document}

\maketitle
\thispagestyle{empty}
\pagestyle{empty}

%%%%%%%%%%%%%%%%%%%%%%%%%%%%%%%%%%%%%%%%%%%%%%%%%%%%%%%%%%%%%%%%%%%%%%%%%%%%%%%%
\begin{abstract}
Many players in the automotive field support scenario-based assessment of automated vehicles (AVs), where individual traffic situations can be tested and, thus, facilitate concluding on the performance of AVs in different situations. 
Since a large number of different scenarios can occur in real-world traffic, the question is how to find a finite set of relevant scenarios. 
Scenarios extracted from large real-world datasets represent real-world traffic since real driving data is used. 
Extracting scenarios, however, is challenging because (1) the scenarios to be tested should assess the AVs behave safely, which conflicts with the fact that the majority of the data contains scenarios that are not interesting from a safety perspective, and (2) extensive data processing is required, which hinders the utilization of large real-world datasets. In this work, we propose an approach for extracting scenarios from real-world driving data.
\cstart The first step is data preprocessing to tackle the errors and noise in real-world data by reconstructing the data. \cend
The second step performs data tagging to label actors' activities, their interactions with each other and the environment.
Finally, the scenarios are extracted by searching for combinations of tags. 
The proposed approach is evaluated using data simulated with CARLA and applied to a part of a large real-world driving dataset, i.e., the Waymo Open Motion Dataset (WOMD). The code and scenarios extracted from WOMD are open to the research community to facilitate the assessment of the automated driving functions in different scenarios\footnote{\url{https://github.com/bDGuo/waymo_motion_scenario_mining.git}}.
\end{abstract}

%%%%%%%%%%%%%%%%%%%%%%%%%%%%%%%%%%%%%%%%%%%%%%%%%%%%%%%%%%%%%%%%%%%%%%%%%%%%%%%%
\section{Introduction}
The development of automated vehicles (AVs) is drawing more and more attention from many industries, including the automotive industry, smart cities, and smart mobility.
AVs have the potential to improve \cstart road capacity and safety \cend since their introduction will gradually reduce crashes resulting from human failure \cite{MilakisSocialImpact}. Although the application of AVs is promising, there are still challenges in the effective validation of AVs to assure \cstart their safe operation \cend in an extremely large number of real-world driving scenarios \cite{Watanabe2019,Song2022}. 

\begin{figure}[tb]
    \centering
    \includegraphics[width=\columnwidth]{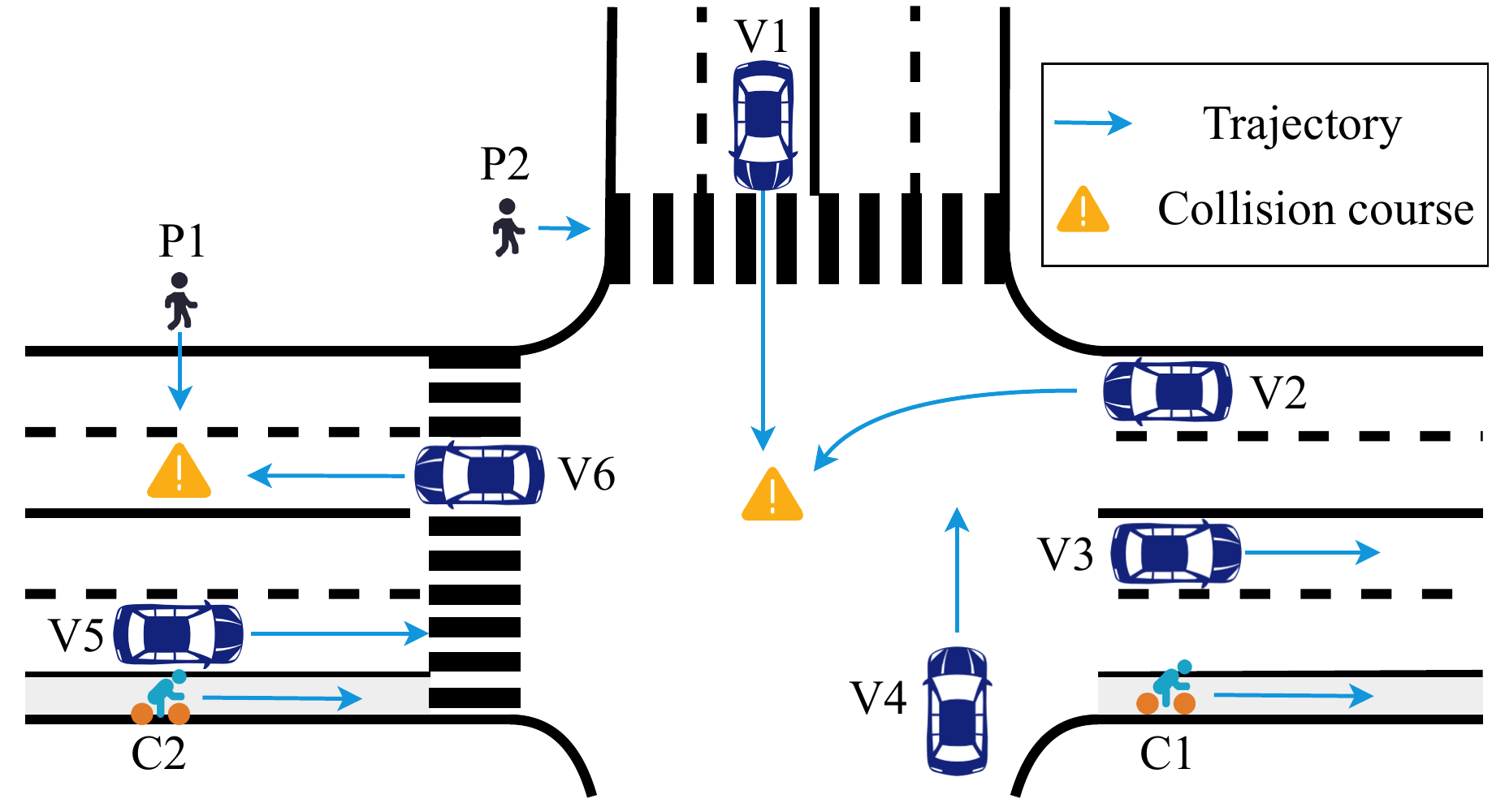}
    \caption{Overview of the traffic where different RU interact with each other and the environment at a crossroads. Vehicles: V1-V6, pedestrians: P1-P2, cyclists: C1-C2.}
    \label{fig:intro}
\end{figure}

 To validate the performance of AVs, \cstart a scenario-based approach has been adopted \cite{ElrofaiScenarioIdentification,Riedmaier2020, Micha2011}.
Comparing with the assessment using data without scenario identification, this approach enables a direct translation of a test result into an assessment of the AV regarding a particular operational design domain (ODD)\cite{manuel2022,DeGelder2022}. \cend
Identified scenarios are crucial for scenario-based assessment since they are directly reflected in the test cases for the assessment~\cite{Stellet2015}. 
To create a common understanding of the scenario-based assessment, definitions for concepts regarding scenarios and their building blocks were proposed in~\cite{DeGelder2022}. Layered models were derived for the scenario description, which contains information on environmental conditions, activities of road users (RU), and their interactions \cite{Menzel2018,Schuldt2017,Scholtes2021}. Previous research regarding scenario identification can be classified as scenario creation based on expert knowledge and scenario extraction from real-world data \cite{CaiSurvey,WangX2022, Hauer2020}. 
A drawback of scenario creation is that the synthesized scenarios may not be  representative of real-world traffic. When using real-world data to extract scenarios, the extracted scenarios are a representation of what an AV might encounter in real-world traffic. The challenge is  the extraction of the relevant, i.e., safety-critical, scenarios from a large amount of data, since the majority of the data contains scenarios that are irrelevant from a safety perspective. Consider  \cref{fig:intro}, where multiple road users interact with each other and the environment at a crossroads.
% Not all interactions in \cref{fig:intro} are relevant for the safety assessment.
The extraction of potentially hazardous interactions between two actors, e.g., V1-V2, and P1-V6, is necessary since they are on collision course, whereas interactions V1-P2 and V2-C1, are not extracted since those interactions are not relevant for the safety assessment.

To be representative of real-world traffic, scenario extraction requires a large amount of data, with which it is possible to identify more uncommon scenarios than with small datasets. Recently released large-scale datasets \cite{CaesarnuScenes, Argo, Ettinger2021, MaAppoloScape} alleviate the effort of collecting data. However, extracting useful information remains a work that requires extensive data processing. The datasets are usually in large file sizes and hinder their utilization~\cite{Zhao2017ITSC,Waymobehavior}.

In this work, we propose an approach to automatically extract scenarios from real-world data with three steps: data preprocessing, automatic tagging, and searching for the combination of tags based on~\cite{GelderRealWorldScenarioMining}. We illustrate the proposed approach by extracting scenarios from a large-scale dataset. The contribution of this work is summarized as follows:
\begin{enumerate}
    % \item We propose a method for mining scenarios from real driving data using automatic tagging and searching for the combination of tags predefined in the scenario category. We tagged the road users' activities, the interaction between road users, and the interaction between the road users and the environment. 
    % \item Our tags include the activities of RU, the interaction between different RU, and the interaction between the RU and the environment, which are suitable for multiple layered models (e.g., \cite{Menzel2018,Schuldt2017,Scholtes2021}) for the scenario description.
    \item In addition to \cite{GelderRealWorldScenarioMining}, our tags include not only the activities of RU but also the interaction between each other and the environment. Our tags are suitable for multiple layered models (e.g., \cite{Menzel2018, Schuldt2017, Scholtes2021}).
    \item We evaluate our method using data simulated with CARLA \cite{CARLA}, where we have access to the ground truth of scenarios, and we can select the type of scenarios generated. Based on the generated data, we verified that our method correctly extracts the scenarios.
    \item We provide an open-source library for the code and scenarios extracted from a large real-world dataset, i.e., the Waymo Open Motion Dataset (WOMD) \cite{Ettinger2021}. \cstart This can facilitate the development and assessment of AVs using data with identified scenarios and enable the translation of a test result into an assessment with respect to the corresponding ODD.\cend
    % \item We facilitate the evaluation of trajectory prediction models on scenario categories, which allows for the evaluation of each scenario category individually  and switching to the best prediction models under certain scenario categories. This approach reveals promising possibilities for the evaluation, such as weighted average errors according to the severity of the encountered traffic scenario.
\end{enumerate}

The remainder of this work is structured as follows. \cref{sec:related_work} introduces related work on scenario extraction. %from real-world driving data. 
\cref{sec:method} presents our scenario extraction methods. 
% \cref{sec:evaluation} describes the evaluation of the proposed approach.
\cref{sec:results} describes the scenarios extracted from simulated data and real-world driving data. This paper ends with conclusions and a discussion.

%%%%%%%%%%%%%%%%%%%%%%%%%%%%%%%%%%%%%%%%%%%%%%%%%%%%%%%%%%%%%%%%%%%%%%%%%%%%%%%%
\section{Related work}\label{sec:related_work}

Scenarios are conceptualized using ontologies in \cite{Geyer2014, Bagschik2018, DeGelder2022}, which define the basic entities and describe relations among them. The basic entities are actors and environmental elements. Actors are the entities that experience change and can act and react in a scenario  \cite{iso34501, DeGelder2022}. Following this definition, we term road users, e.g., vehicles, cyclists, and pedestrians, as actors. We adopt the term activity from \cite{DeGelder2022}, which describes the state change that actors experience in a certain time interval. Environmental elements entail road networks, such as crosswalks, and traffic lights. The relations among the basic entities include the interaction between the actors and the environmental elements and the interaction between different actors.

Next, a selection of previous publications on extracting scenarios from real-world data is introduced. In \cite{Erwin2017} the importance of using scenarios extracted from real-world driving data is highlighted since it allows drawing conclusions on the performance of AVs in real-world traffic.
%De Gelder and Paardekooper \cite{Erwin2017} highlight the importance of using scenarios extracted from real-world driving data, since it allows drawing conclusions on the performance of AVs in real-world traffic. Therefore, this paper focuses on extracting scenarios from real-world driving data.
% \subsection{Ontology for scenario-based assessment}

% Bagschik et al.\ \cite{Bagschik2018} proposed a layered approach to generate traffic scenarios based on expert knowledge. However, experts are unlikely to identify all scenarios. This study aims at mining scenarios from a real-world database to provide a wide range of scenarios.

% In \cite{DeGelder2022}, the definition of scenario is developed. Scenarios and their elementary blocks are defined as classes of objects in an object-oriented framework (OOF). We adopt the definitions of scenarios and all of their essential building blocks (such as activities, actors, and events) in \cite{DeGelder2022} for clarity.

% \subsection{Real-world scenario extraction}

% Different techniques for scenario extraction on highways are proposed in the literature.
In \cite{Karunakaran2022HighD}, lane change scenarios are extracted by comparing vehicles' lane position and lateral distance to the path of the ego vehicle from real-world data. The lane points are clustered into different lanes. Lane changes are detected if any vehicle's lateral displacement to the ego vehicle's lane is less than a threshold.

In \cite{ElrofaiScenarioIdentification}, detection for turning scenarios is discussed.
% The yaw rate is estimated with the rear wheel velocities and the wheel track since the authors argued that the available yaw rate sensor was inaccurate. 
The fluctuations of the estimated yaw rate are smoothed out with an exponentially decaying weighted averaging filter. The turning scenario is detected when the filtered signal crosses a threshold for a certain time interval. This approach, however, can miss quick turns and slow turns where this condition does not hold.
% This work reveals the importance of data preprocessing. 

In \cite{GelderRealWorldScenarioMining}, scenarios are extracted from real-world driving data by labeling the data with tags, e.g., the lateral and longitudinal activities of different actors, and searching for a combination of tags. This approach is efficient since the tags are only detected once and shared while searching for a combination of tags to extract different types of scenarios. The authors determined the longitudinal activities of the actors on highways by looking at the speed difference in a certain sample window.
% The interaction between the ego vehicle and the environment is determined by the location of the ego vehicle based on GPS measurements.
We extend this approach with longitudinal tags that are suitable for the urban area.
% e.g., standing still and reversing.

%%%%%%%%%%%%%%%%%%%%%%%%%%%%%%%%%%%%%%%%%%%%%%%%%%%%%%%%%%%%%%%%%%%%%%%%%%%%%%%%
\section{Methodology}\label{sec:method}
\begin{figure*}[htbp]
    \centering
    \includegraphics[width=\textwidth]{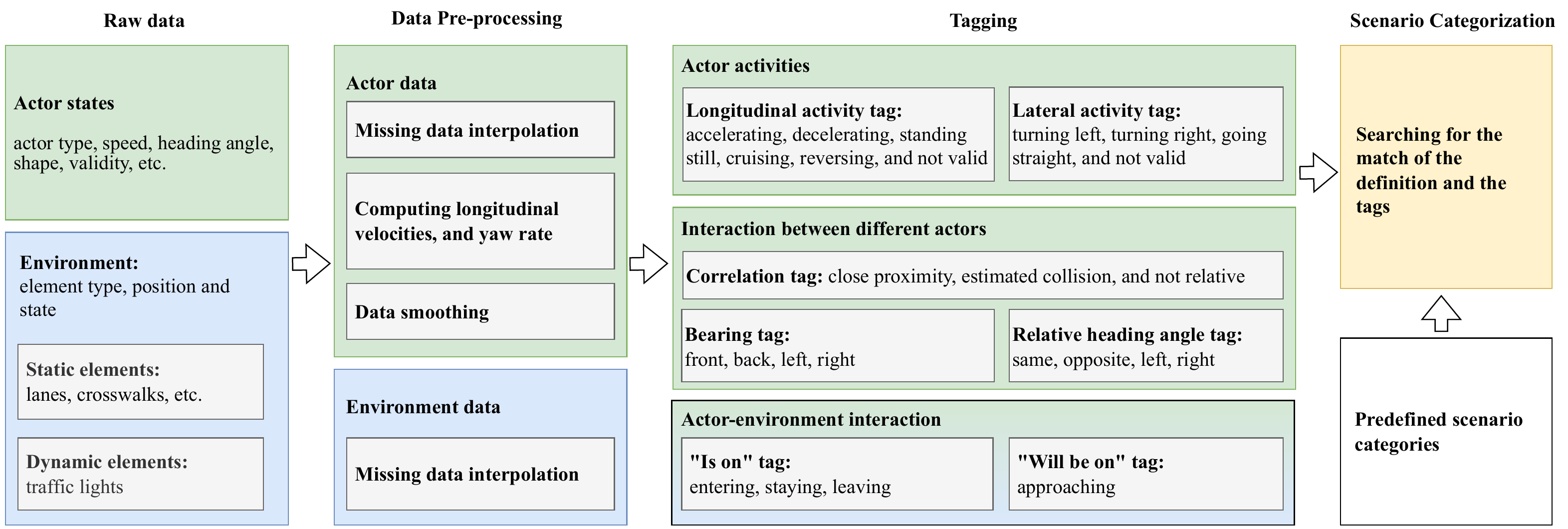}
    \caption{The three-step approach for scenario extraction from real-world data. The colors green and blue refer to actor-related and environment-related processes and tags, respectively.}
    \label{fig:proposed_method_overview}
\end{figure*}

In this section, the approach for real-world scenario extraction is introduced. As shown in \cref{fig:proposed_method_overview}, our approach is captured into three steps: data preprocessing, tagging, and scenario categorization. The first step, data preprocessing, addresses inaccuracies in the data. The next step, tagging, is completed by a model-based approach to tag the longitudinal and lateral activities of the actors and their interaction with the environment and each other.
The third step, scenario categorization, is done by searching for a combination of tags. In the following subsections, these three steps are detailed. We assume that the data is in a coordinate frame with the $x$-axis pointing east and the $y$-axis pointing north.

\subsection{Data preprocessing}
Despite the benefits of large datasets, noise and missing data in these datasets are inevitable which influences the reliability of future research \cite{Waymobehavior}.
\cstart To tackle the errors, we reconstruct the trajectories of the actors with the assumption that an unsmooth longitudinal velocity or missing measurement is a measurement error. \cend
%The missing data can stem from a sensor error or an object that is occluded for a certain time interval. 
To counter this, we linearly interpolate the missing data between the first and the last valid time step. 
The actor's yaw angle is denoted by $\psi \in (-\pi,\pi]$. 
We compute the yaw rate, $\omega$, as 
\begin{equation}\label{eq:omega}
        \omega(k+1)=\left(\psi(k+1)-\psi(k)\right) /\ \timestep,
    \end{equation}
where $\psi(k)$ denotes the yaw angle at time step $k$ and $\timestep$ denotes the sampling time.
    
% $\Delta_{\psi}(k+1)=\psi(k+1)-\psi(k)$ 
%In case $\psi$ changes from $\pi$ to $-\pi$, or vice versa, we add or subtract $2\pi\timestep$ from $\omega$ such that $\omega\in(-\pi\timestep, \pi\timestep]$, based on the assumption that the yaw angle change of an actor in a time step is physically restrained. 
Next, we compute the actor's longitudinal velocity $v_{\mathrm{long}}$:
\begin{equation}\label{eq:v_long}
    v_\mathrm{long} = \cos{\psi}\ v_x + \sin{\psi}\ v_y,
\end{equation}
where $v_x$ and $v_y$ denote the actor's velocities along the $x$-axis and the $y$-axis, respectively. We utilize cubic splining \cite{deboor1978practical} to smooth the longitudinal velocity.
    % \begin{equation}\label{eq:dot_psi2}
    %     \Delta_{\psi}(t) = \begin{cases}
    %     \Delta_{\psi}(t) & |\Delta_{\psi}(t)| < \pi \\
    %     sgn(\Delta_{\psi}(t))\cdot (|\Delta_{\psi}(t)|-2\pi) & |\Delta_{\psi}(t)| \geq \pi
    %     \end{cases}
    % \end{equation}
    % \begin{equation}\label{eq:dot_psi3}
    %     \omega(t) = \begin{cases}
    %         0 & t=0 \\
    %         (\Delta_{\psi}(t) - \Delta_{\psi}(t-1)) \cdot T_s & t>0
    %     \end{cases}
    % \end{equation}
    
\subsection{Tagging}\label{sec:tagging}
To compose the scenario, we introduce three tag classes: actor activities, actor-environment interaction, and interaction between different actors. Actor activities entail actors' longitudinal and lateral activities, further explained in \cref{subsec:actor_activity}. Actor-environment interaction describes how each actor interacts with the environment and is detailed in \cref{subsec:relateEnvionmentActor}. The interaction between different actors in \cref{subsec:inter-actor} comprises the potential hazardous interaction type of two actors. The tag ``not valid'', which represents the time sequence before the first valid data and after the last valid data, is included in three tag classes.

\subsubsection{Actor activity}\label{subsec:actor_activity}

Five different types of longitudinal activity tags are distinguished, i.e., ``accelerating'', ``decelerating'', ``standing still'', ``cruising'', and ``reversing''. %An actor is performing either one of these activities. 
We use the method in \cite{GelderRealWorldScenarioMining} for the tags ``accelerating'', ``decelerating'', and ``cruising''.
% This method uses the speed difference within a certain sample window of length to generate tags: ``accelerating''}, ``decelerating''}, and ``cruising''}. We extended the method with two tags: ``standing still''} and ``reversing''}.
The following rules are used to tag the activities ``reversing'' or ``standing still'', respectively:
% An implementation of this algorithm is shown in \cref{fig:long_result}.
%% TODO: put the equation in text to shorten it.
\begin{equation}\label{eq:reversing}
    v_{\mathrm{long}}(k)\ \timestep < -\alpha\ l_{\mathrm{actor}},
\end{equation}
\begin{equation}\label{eq:standingstill}
    |v_{\mathrm{long}}(k)|\ \timestep \le \alpha\ l_{\mathrm{actor}},
\end{equation}
where $v_{\mathrm{long}}$ is the longitudinal velocity, $l_{\mathrm{actor}}$ is the length of the actor's bounding box, and $\alpha\in(0,1)$ is a tuning parameter.% that needs to be set in advance. 
In this work, we use $\alpha=0.01$.

Lateral activity tags are distinguished into three types, which are ``turning left'', ``turning right'', and ``going straight''. To extract lateral activities, simply checking whether the yaw rate $\omega$ is above a certain threshold $\lambda_{\omega}$ for a certain time span would miss quick and slow turns.
To tackle this problem, we introduce a threshold $\lambda_{\psi}$ for future heading change. Below the method is discussed for "turning left".
%Following is the method for detecting ``turning left'' by using the thresholds $\lambda_\omega$ and $\lambda_\psi$. 
\begin{enumerate}
    % \item Assume the actor is going straight. \textcolor{red}{remove?}
    \item 
    % Let $\lambda_{\omega}$ denote the threshold for yaw rate $\omega$.
    The current time step $\currenttimestep$ is staged as the potential start of ``turning left'' if %the yaw rate is more than $\lambda_{\omega}$:
\begin{equation}\label{eq:yawratesamplingthreshold}
        \omega (\currenttimestep) > \lambda_{\omega}.
    \end{equation}
    \item Determine the nearest time step $\timestepend$ of this lateral activity %when the yaw rate is less than $\lambda_{\omega}$, i.e.,
    \begin{equation}\label{eq:yawrateend1}
        \omega (\timestepend) < \lambda_{\omega}.
    \end{equation}
    \item Compare the heading change with $\lambda_{\psi}$. The lateral activity between $\currenttimestep$ and $\timestepend$ is tagged  ``turning left'' if 
    \begin{equation}\label{eq:turningleftcondition}\timestep\sum_{k=\currenttimestep}^{\timestepend}\omega(k) > \lambda_{\psi}.
    \end{equation}
    % \item If the condition \eqref{eq:turningleftcondition} does not hold, the tag ``turning left" at $k_c$ is canceled. The next iteration checks the time step $k_c+1$ with \cref{eq:yawratesamplingthreshold}. \textcolor{red}{remove?}
\end{enumerate}

We set the threshold for the heading change $\lambda_{\psi}=45^{\circ}$ based on \cite{Pool2017}.
Let $\timeintervalturning$ denote the longest time interval for a turning activity.
Therefore, we define $\lambda_{\omega} = \lambda_{\psi} / \timeintervalturning$. The tag ``turning right'' is detected in a similar manner.
% as ``turning left''.

\subsubsection{Actor-environment interaction}\label{subsec:relateEnvionmentActor}
The interaction between actors and environmental elements is distinguished into five types: ``not relative'',  ``approaching'', ``entering'', ``staying'', and ``leaving''. For example, the interaction between a crosswalk and a pedestrian that will possibly be on the crosswalk, e.g., P2 in \cref{fig:intro}, is tagged with ``approaching''. This can be vital for the vehicle on the crosswalk, e.g., V1 in \cref{fig:intro}, where the interaction of V1 to the crosswalk is subsequently tagged with ``entering'', ``staying'', and ``leaving''. An actor can only possesses one tag with one environmental element at each time step. The environmental elements are commonly provided with coordinates of polylines, e.g., lane center, and vertices of polygons, e.g., crosswalks \cite{Argo,Ettinger2021}. To detect the tags, we formulate the environmental elements in polygons. We introduce the intersection ratio $\phi_a$ to detect if the actor is ``on'' the corresponding environmental element. The intersection ratio is obtained from the intersection area $\mathcal{A}_\mathrm{in}$ of the actor’s bounding box and the environmental element polygon normalized by the area of the actor's bounding box $\mathcal{A}_\mathrm{actor}$,
\begin{equation}\label{eq:eq_AIR}
    \phi_a = \mathcal{A}_\mathrm{in} / \mathcal{A}_\mathrm{actor}.
\end{equation}
The normalization facilitates the adaptation of $\phi_a$ to actors of different sizes. The indicator used to distinguish the status ``on'' to tags: ``staying'', ``entering'', and ``leaving'' is the difference of the actual intersection ratio $\Delta\phi_a$ obtained with,
\begin{equation}\label{eq:eq_airdifference}
    \Delta\phi_a=\phi_a(k+1)-\phi_a(k).
\end{equation}
The tag ``approaching'' is detected by introducing the extended trajectory polygon $\mathcal{P}_e$ and the extended intersection ratio $\phi_e$. We extend the current actor's bounding box to the extended trajectory polygon with a constant turn rate and velocity model (CTRV) \cite{Schubert2011} in order to extract the actor's intention to move toward a certain environmental element within a time horizon $\extendinghorizon=\SI{3}{\second}$. The method for generating $\mathcal{P}_e$ is as follows:
% \textcolor{red}{perhaps pseudo code will be better to illustrate how to generate ETP with a constant velocity and yaw rate model}

\begin{enumerate}
    \item At each time step in $\extendinghorizon$, a sequence of polygons is generated with CTRV.
    % \begin{equation}\label{eq:cvyr}
    %     \begin{cases}
    %         \psi(t+1) = \psi(t) + \omega(t) \cdot T_s \\
    %         \begin{aligned}
    %             x(t+1) = x(t) + T_s \cdot v(t) \cdot (sin(\psi(t+1))-\\sin(\psi(t)))
    %         \end{aligned}
    %         \\
    %         \begin{aligned}
    %             y(t+1) = y(t) - T_s \cdot v(t) \cdot (cos(\psi(t+1))+\\cos(\psi(t)))
    %         \end{aligned}
    %     \end{cases},
    % \end{equation}
    % where $(x,y)$ is the coordinate of polygon center.
    \item An extended trajectory polygon is the union of the sequence of polygons in the first step.
\end{enumerate}
The extended intersection ratio $\phi_e$ is the intersection area of $\mathcal{P}_e$ and the environmental element normalized by the area of $\mathcal{P}_e$.
The rules for generating actor-environment interaction tags are summarized in \cref{tab:static-actor}.
If both $\phi_a = 0$ and $\phi_e = 0$, the actor is not interacting with the environmental element.
%The tag ``approaching'' is used if $\phi_a = 0$ and $\phi_e > 0$. 
%If $\phi_a > 0$, one of the tags ``staying'', ``entering'', or ``leaving'' is applicable, depending on the value of $\Delta \phi_a$: ``entering'' if $\Delta\phi_a > 0.01$, ``leaving'' if $\Delta\phi_a < -0.01$, and ``staying'' otherwise.

\begin{table}[tb]
	\centering
	\caption{Rules for generating actor-environment interaction tags. $\phi_a$: actual intersection ratio of \cref{eq:eq_AIR}, $\phi_e$: extended intersection ratio, $\Delta\phi_a$: the difference of $\phi_a$ of \cref{eq:eq_airdifference}.}
	\label{tab:static-actor}
	\begin{tabular}{cccc}
		\toprule
		$\phi_a$ &  $\phi_e$ & $\Delta\phi_a$ & Tag\\ \midrule
        $\phi_a=0$ & $\phi_e=0$ & - & Not relative  \\
		$\phi_a=0$ & $\phi_e>0$ & - & Approaching \\
        $\phi_a>0$ & - & $\left| \Delta\phi_a\right| \leq 0.01$ & Staying \\
        $\phi_a>0$  & - & $\Delta\phi_a > 0.01$ & Entering \\
        $\phi_a>0$  & - & $\Delta\phi_a < -0.01$  & Leaving\\ \bottomrule
	\end{tabular}
\end{table}

\subsubsection{Interaction between different actors}\label{subsec:inter-actor}

\begin{figure}[tb]
    \centering
    \includegraphics[width=0.45\columnwidth]{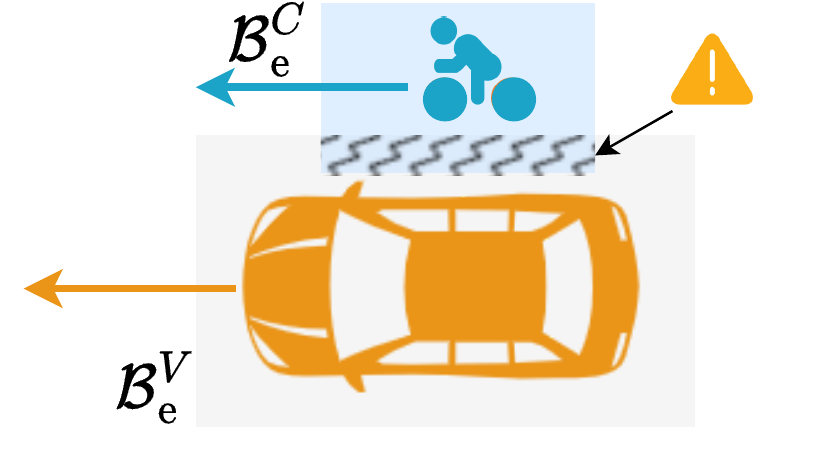}
    \caption{The expanded bounding boxes $\mcBe^V$ and $\mcBe^C$ for tagging the interactive actors in close proximity. $C$: cyclist, $V$: vehicle.}
    \label{fig:interactor-a}
\end{figure}

The tag class describing the interaction between different actors is designed based on two observations. The first observation is that actors do not react to every actor in the field of view; rather, they selectively focus on the key actors whose movements will (potentially) collide with theirs. We use the tag ``estimated collision'' to label this type of interaction. 
The second observation stems from the collision hazard with the actors in a nearby neighborhood and the heterogeneity of different actors. For instance, a vehicle and a pedestrian keep a safe distance from each other to avoid a collision while passing by, where the car's safe distance is considered larger than that of the pedestrian due to the larger shape of the car. 
We use the tag ``close proximity'' to label this type of interaction. 
%We term actors tagged with either ``close proximity'' or ``estimated collision'' as interactive actors.
% and the tags of their bearing angle and the relative heading angle are appended.
% We distinguished the near-collision tag into three types:  ``close proximity'', ``estimated collision'', and ``not relative''.
%The tag ``close proximity'' 
This tag is generated by introducing the expanded bounding boxes to adapt to diverse actors' shapes. Let $\mcBe^i$  denote the expanded bounding box of actor $i$.
% describes the instantaneous bounding boxes interaction of the actors. 
% We may check whether the shortest distance between two actors' bounding boxes is less than a safe distance. However, this approach is not adaptive to actors of different extents. For example, the safe distance of a car is considered as larger than that of a pedestrian. Instead of setting a safety distance, 
To generate $\mcBe^i$, the length and width of the original actor's bounding box are multiplied with a parameter $\beta=2$ for expansion. 
The interaction between actor $i$ and $j$ is tagged with ``close proximity'' when $\mcBe^i$ and $\mcBe^j$ intersect, e.g., \cref{fig:interactor-a}. 

\begin{figure}[tb]
    \centering
    \includegraphics[width=0.7\columnwidth]{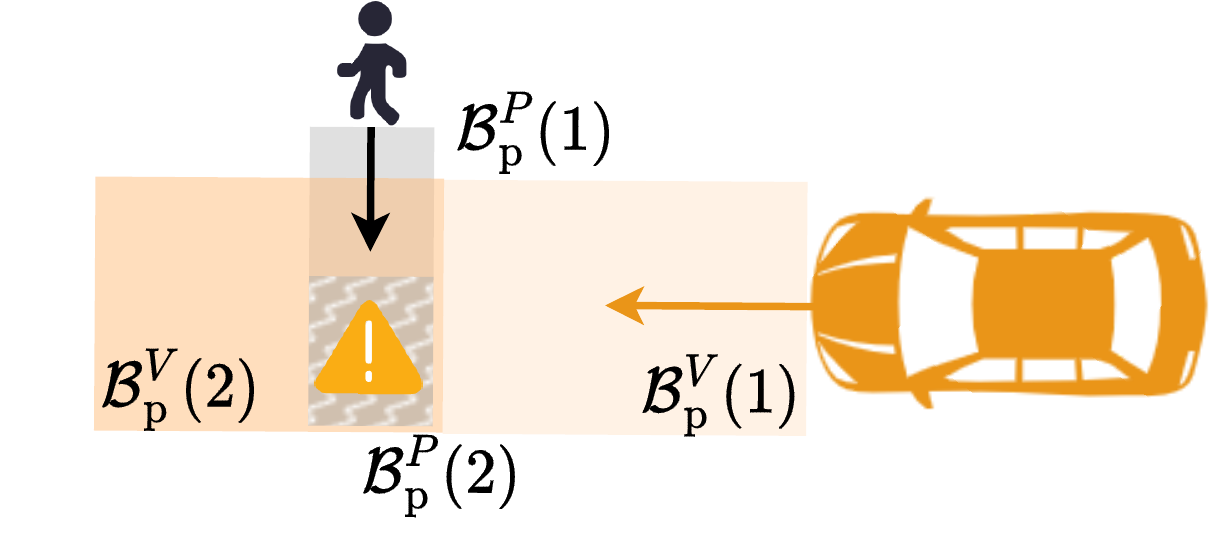}
    \caption{The predicted bounding boxes $\mcBp^P$ and $\mcBp^V$ for tagging the interactive actors with estimated collision. $P$: pedestrian, $V$: vehicle, $\mcBp(\timesteppredict)$: $\mcBp$ at the time step $\timesteppredict$.}
    \label{fig:interactor-b}
\end{figure}

The tag ``estimated collision'' is given by introducing the predicted bounding boxes $\mcBp^i$ for an actor $i$. At each time instance, $\mcBp^i$ is a sequence of polygons generated with CTRV \cite{Schubert2011}, which outputs a time series of the future bounding boxes of the $i$-th actor in a prediction time horizon. The actors are tagged with ``estimated collision'', e.g., \cref{fig:interactor-b}, when the predicted bounding boxes in any time step of the prediction time horizon overlap with each other,
\begin{equation}\label{eq:PBB}
        \mcBp^i(\timesteppredict) \cap \mcBp^j(\timesteppredict) \neq \O, \exists\ \timesteppredict \in \{1,\ldots,\predictionhorizon / \timestep\},
\end{equation}
%where $\mcBp^i(\timesteppredict)$ and $\mcBp^j(\timesteppredict)$ are the predicted bounding boxes of two different actors at the time step $\timesteppredict$ and 
where $\predictionhorizon$ denotes the prediction time horizon. In this work, we use $\predictionhorizon=\SI{5}{\second}$. The tag ``not relative'' is given to two interactive actors at the time instances not tagged with ``close proximity'' or ``estimated collision''. In the remainder of this work, we term one of two interactive actors as the host actor and the other as the guest actor. %Note that the selection of the host and the guest actor is arbitrary.


% \begin{figure}
%     \centering
%     \subfloat[The EBB for tagging the interactive actors in close proximity.\label{fig:interactor-a}]{
%     \includegraphics[width=0.45\columnwidth]{figures/EBB.pdf}}\\
%     \subfloat[The PBB for tagging the interactive actors with estimated collision. \label{fig:interactor-b}]{\includegraphics[width=0.78\columnwidth]{figures/PBB.pdf}}
%     \caption{Examples for the usage of EBB and PBB. (C: cyclist, V: vehicle, P: pedestrian, PBB(t): the PBB at the time instance $t$)}
%     \label{fig:interactor}
% \end{figure}

 To describe the interaction of different actors, we further tagged the interactive actors with their relative heading and bearing angles. The relative heading angle is the angle required to rotate the heading vector of the host actor to that of the guest actor. The bearing angle is the angle to rotate the heading vector of the host actor to the bearing vector, which points from the center of the host to that of the guest actor.
 % \textcolor{red}{Replace the blue text in this part with an illustrative figure ?}
 The rules for tagging the relative heading angle and the bearing angle are provided in \cref{tab:inter-actor-heading-pos}.
\begin{table}%[tb]
     \centering
	\caption{Rules of tagging relative heading and bearing angles of actors interacting with each other.}
	\label{tab:inter-actor-heading-pos}
	\begin{tabular}{ccc}
\specialrule{0em}{3pt}{3pt}
  \toprule
     Angle range / \SI{}{\radian} &  Relative heading tag & Bearing tag \\ \midrule
    (-$\pi$,-0.75$\pi$] or (0.75$\pi$,$\pi$] & Opposite & Back \\
  (-0.75$\pi$,-0.25$\pi$] & Right & Right \\
  (-0.25$\pi$,0.25$\pi$] & Same & Front \\
  (0.25$\pi$,0.75$\pi$] & Left & Left\\ \bottomrule
	\end{tabular}
\end{table}

\subsection{Scenario categorization using tags}
Scenarios are categorized using a combination of tags. For example, \cref{tab:exp-sc} shows the definition matrix of the scenario category ``vehicle-to-cyclist passing by''. In the scenarios that fall in this scenario category, a vehicle and a cyclist are going straight while passing each other closely, and the cyclist is on the left or right side of the vehicle. The scenarios are extracted by searching for matches within the tags\cite{GelderRealWorldScenarioMining}.%, where the logical rules AND, OR, or NOT may be used \cite{GelderRealWorldScenarioMining}.  

\begin{table}%[tb]
\centering
\caption{Definition matrix of the scenario category ``vehicle-to-cyclist passing by''.}
\label{tab:exp-sc}
\resizebox{\columnwidth}{!}{%
\begin{tabular}{p{0.9in}p{1in}p{1in}}
\specialrule{0em}{3pt}{3pt}
\toprule
                               & Host actor tag     & Guest actor tag\\ \midrule
Actor type                       & Vehicle         & Cyclist                          \\ \hline
Longitudinal activity & Accelerating OR decelerating OR cruising & Accelerating OR decelerating OR cruising \\ \hline
Lateral activity                 & Going straight  & Going straight                   \\ \hline
Interaction with the environment & -               & -                                \\ \hline
Interaction between actors            & Close proximity & -                                \\ \hline
Bearing                          & Left OR right     & -                                \\ \hline
Relative heading                 & Same            & -                                \\ \bottomrule
\end{tabular}%
}
\end{table}

%%%%%%%%%%%%%%%%%%%%%%%%%%%%%%%%%%%%%%%%%%%%%%%%%%%%%%%%%%%%%%%%%%%%%%%%%%%%%%%%
\section{Results}\label{sec:results}
To evaluate our method, the algorithm is first analysed with data generated with the CARLA simulator \cite{CARLA}.
Previous studies \cite{ElrofaiScenarioIdentification, GelderRealWorldScenarioMining} evaluated their approaches with manually labeled real-world driving data, which is not feasible for large datasets.
By using simulated data to evaluate the scenarios, we can customize the environment conditions and traffic flow of the data, where we have access to the ground truth of scenarios, and we can select the type of scenarios generated.
To evaluate the performance of our approach, we customize the three scenario categories shown in \cref{tab:evaluation} and record 30 simulations in different areas. %, e.g., surface streets, crosswalks, and intersections, where a vehicle interacts with different types of actors.
For this experiment, we have used $\timestep=\SI{0.05}{\second}$ and $\timeintervalturning=\SI{20}{\second}$.
The proposed algorithm correctly extracts all of the three predefined scenario categories, and no scenarios were incorrectly extracted. 

\begin{table}[bt]
\centering
\caption{Scenario categories for evaluation and scenario mining.}
\label{tab:evaluation}
{%
\begin{tabular}{p{0.18in}p{2.82in}}
\specialrule{0em}{3pt}{3pt}
\toprule
  & Description                                                          \\ \midrule
SC1 & A vehicle is turning left. Another vehicle is going straight in the opposite direction. The two vehicles are on collision course. \\
SC2 & A vehicle and a cyclist are going straight while passing each other closely. The cyclist is on either side of the vehicle.    \\
SC3 & A pedestrian is crossing a vehicle's lane and the pedestrian and vehicle are on collision course. \\ \bottomrule
\end{tabular}%
}
\end{table}

\begin{figure*}[tb]
    \centering
    \subfloat[SC1\label{fig:resultsWOMD1}]{%
    \includegraphics[width=0.273\textwidth]{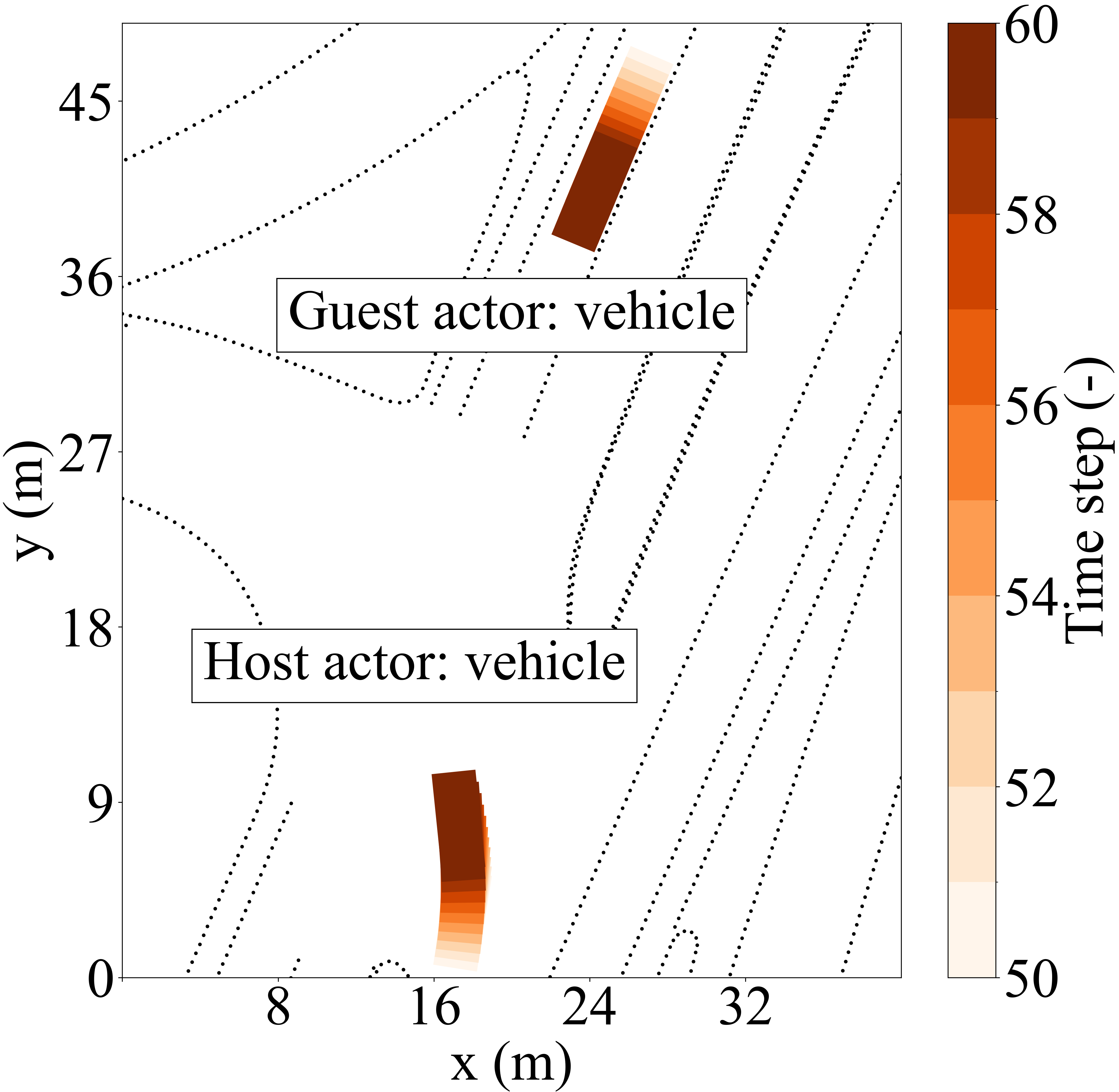}}
    \hfill
    \subfloat[SC2\label{fig:resultsWOMD2}]{%
    \includegraphics[width=0.273\textwidth]{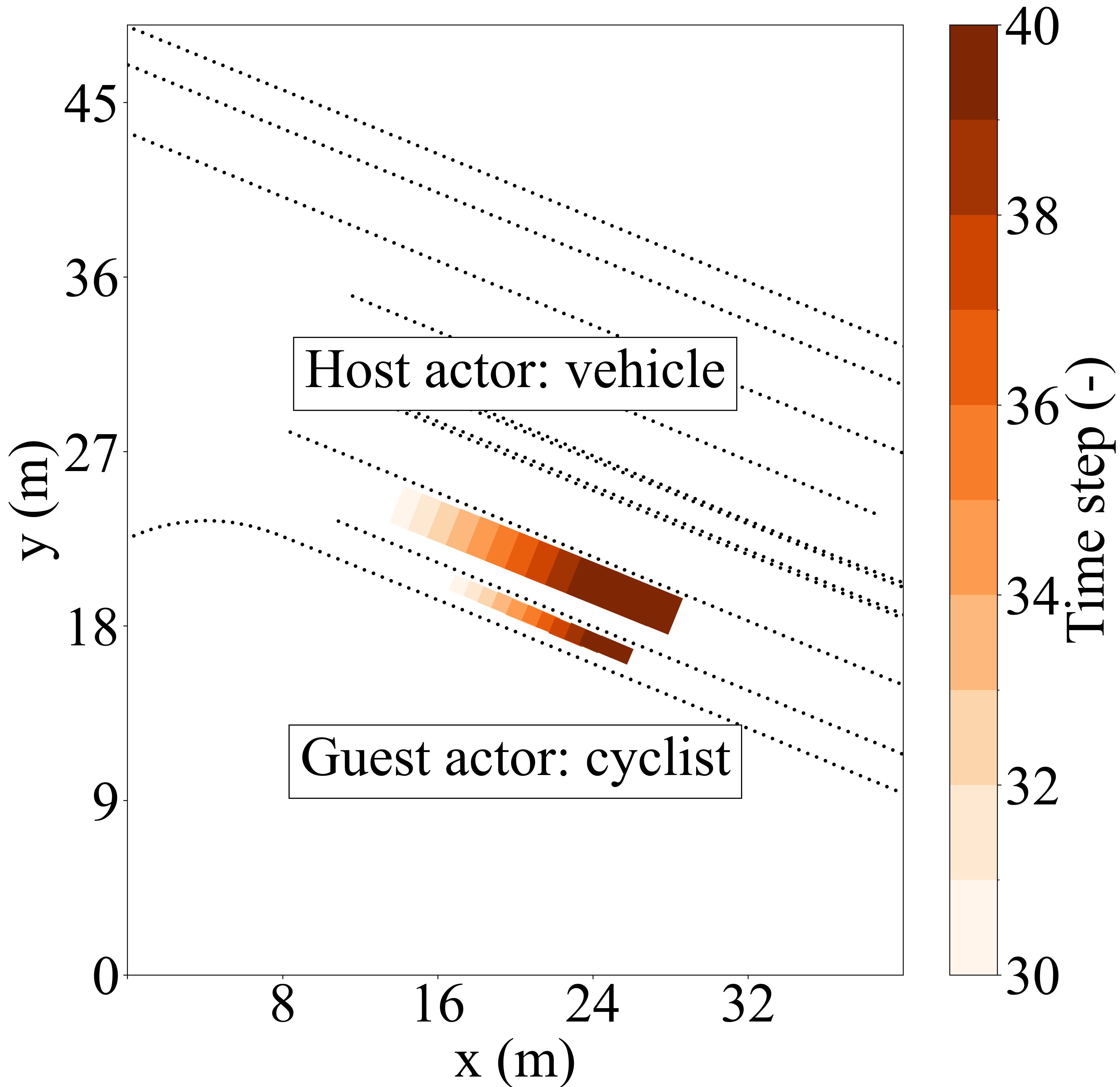}}
    \hfill
    \subfloat[SC3\label{fig:resultsWOMD3}]{%
    \includegraphics[width=0.305\textwidth]{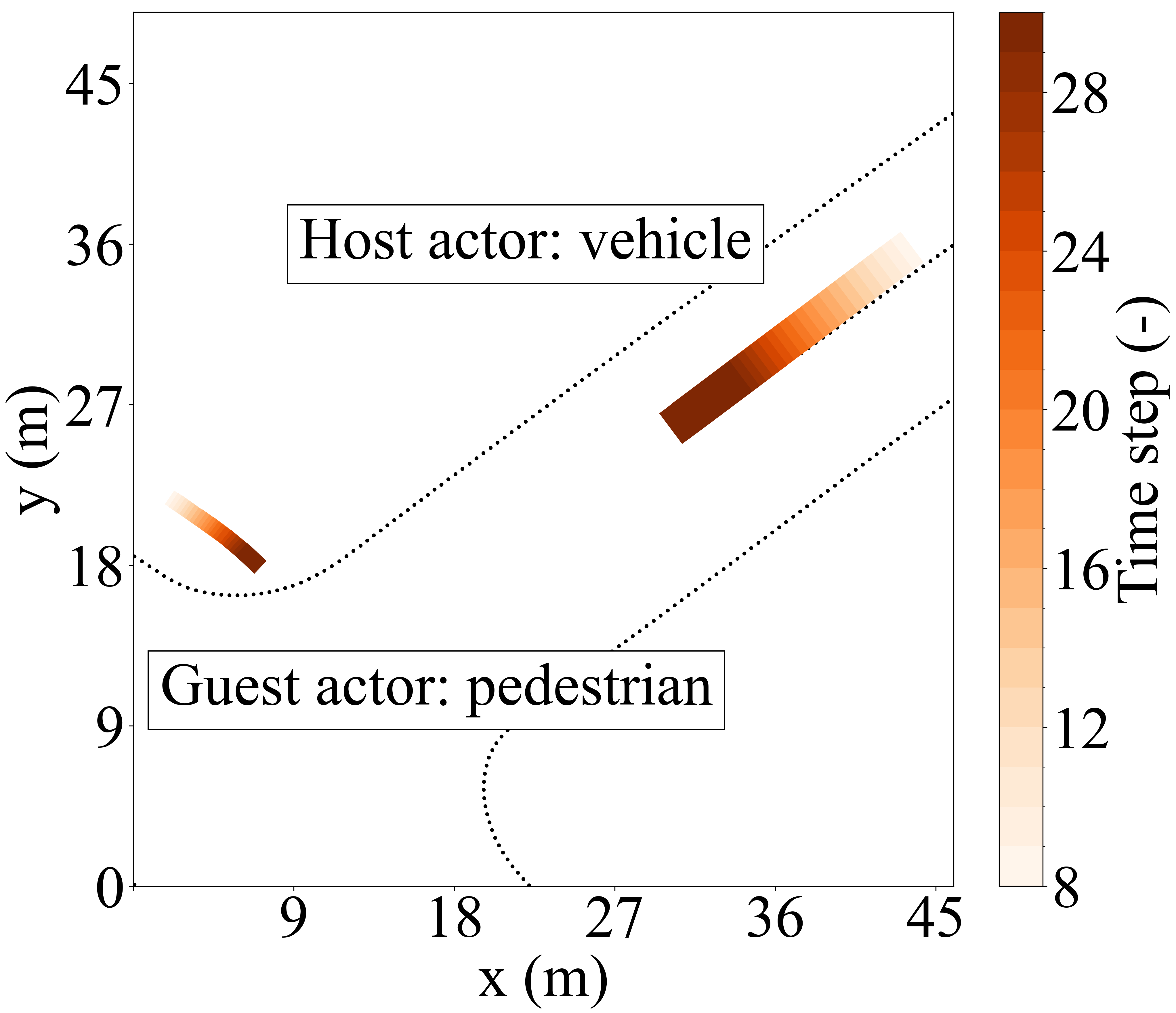}}
    \caption{Examples for the extracted scenarios. The origin of the coordinate system is an arbitrary point.}
    \label{fig:resultsWOMD}
\end{figure*}

To illustrate the applicability, our approach is applied to the training subset WOMD \cite{Ettinger2021}. The training subset includes 1,000 data sequences.
Each data sequence contains \SI{9.1}{\second} of real-world traffic sampled at \SI{10}{\hertz}, thus we use $\timestep=\SI{0.1}{\second}$ and $\timeintervalturning=\SI{9.1}{\second}$.
The total data used in this paper contains approximately 8.68 million vehicle trajectories, 0.97 million pedestrian trajectories, and 7.78 thousand cyclist trajectories. The considered data contains approximately 320 million vehicle-vehicle interactions, 63.85 million vehicle-pedestrian interactions, and 4.34 million vehicle-cyclist interactions. The three types of scenarios listed in \cref{tab:evaluation} are extracted, which in total consist of 215,090 scenarios. To assess the performance, 116 scenarios were randomly extracted and visualized, see \cref{fig:resultsWOMD} for an example. No examined scenarios are incorrectly extracted. There are approximately \SI{50}{\percent}, \SI{12}{\percent}, and \SI{38}{\percent} of the extracted scenarios that fall in the scenario category SC1, SC2, and SC3, respectively.

%A challenge with large datasets is that the majority of the data typically contains scenarios that are not interesting from a safety perspective. 
%This is illustrated by the fact that the considered data contains approximately 320 million vehicle-vehicle interactions, 63.85 million vehicle-pedestrian interactions, and 4.34 million vehicle-cyclist interactions.
%Only a small part of these interactions are considered for the three types of scenarios listed in \cref{tab:evaluation}.

As a start toward building a real-world scenario database with more scenario categories, we observe an imbalance in the distribution of the three scenario categories. This imbalance can result in misjudgement of data-driven models for AVs, which are often evaluated by averaging errors over large datasets without scenario distinction \cite{manuel2022}. For instance, consider two trajectory prediction models, A and B. Model A is more accurate in SC1, and model B is more accurate in SC2. Given the greater prevalence of SC1 scenarios in the dataset, model A would be deemed
superior \cstart when the models are evaluated using data without scenario distinction. \cend However, in datasets with many cyclists, model B would be preferred. Therefore, it is crucial to \cstart distinguish the real-world data in scenario categories and \cend consider the ODD where the models will be deployed and to assess which models are best suited for each specific domain.% according to the evaluations performed on the corresponding scenario categories.

We provide an open-source library for the code and the extracted scenarios. The scenarios come with corresponding sample time, scene ID, and actor ID, which align with WOMD, to facilitate the research community to track and understand this large real-world driving dataset.

%%%%%%%%%%%%%%%%%%%%%%%%%%%%%%%%%%%%%%%%%%%%%%%%%%%%%%%%%%%%%%%%%%%%%%%%%%%%%%%%
\section{Conclusion and future work}\label{sec:conclusion_future}
% \textcolor{red}{TODO}
Scenario identification is crucial for the safety assessment of automated vehicles. In this work, a three-step approach for extracting scenarios from large real-driving datasets is proposed.
The first step involves data preprocessing to fill in missing data and reduce noise. The second step performs data tagging. There are three classes of tags: actor activity, actor-environment interaction, and interaction between different actors. The last step extracts the scenarios by searching for a combination of tags.

The approach is evaluated with the data simulated with CARLA \cite{CARLA}, %where the traffic and the environmental conditions can be customized. 
By using simulated data, the ground truth of scenarios is accessible allowing for accurate assessment of the algorithm.
The algorithm is applied to the training subset of a large real-world driving dataset, i.e., WOMD \cite{Ettinger2021}. A total of 215,090 scenarios across three scenario categories were extracted. %There is an imbalance in the distribution of the three scenario categories that can result in misjudgment while evaluating data-driven models, where errors are often computed equally among different types of scenarios \cstart if there is no scenario identification of the data. 
\cend The code and the extracted scenarios are open to the research community to gain an understanding of the scenarios contained in WOMD. Future work includes labeling WOMD with more tags and extracting more scenario categories to build a real-world scenario database.

% %%%%%%%%% REFERENCES
\bibliographystyle{IEEEtran}
\bibliography{IEEEabrv,references}

\begin{thebibliography}{10}
\providecommand{\url}[1]{#1}
\csname url@rmstyle\endcsname
\providecommand{\newblock}{\relax}
\providecommand{\bibinfo}[2]{#2}
\providecommand\BIBentrySTDinterwordspacing{\spaceskip=0pt\relax}
\providecommand\BIBentryALTinterwordstretchfactor{4}
\providecommand\BIBentryALTinterwordspacing{\spaceskip=\fontdimen2\font plus
\BIBentryALTinterwordstretchfactor\fontdimen3\font minus
  \fontdimen4\font\relax}
\providecommand\BIBforeignlanguage[2]{{%
\expandafter\ifx\csname l@#1\endcsname\relax
\typeout{** WARNING: IEEEtran.bst: No hyphenation pattern has been}%
\typeout{** loaded for the language `#1'. Using the pattern for}%
\typeout{** the default language instead.}%
\else
\language=\csname l@#1\endcsname
\fi
#2}}

\bibitem{MilakisSocialImpact}
D.~Milakis, B.~Van~Arem, and B.~Van~Wee, ``{Policy and society related
  implications of automated driving: A review of literature and directions for
  future research},'' \emph{Journal of Intelligent Transportation Systems},
  vol.~21, no.~4, pp. 324--348, 2017.

\bibitem{Watanabe2019}
H.~Watanabe, L.~Tobisch, J.~Rost, J.~Wallner, and G.~Prokop, ``{Scenario Mining
  for Development of Predictive Safety Functions},'' in \emph{2019 IEEE
  International Conference on Vehicular Electronics and Safety (ICVES)}, 2019,
  pp. 1--7.

\bibitem{Song2022}
Q.~Song, P.~Runeson, and S.~Persson, ``{A Scenario Distribution Model for
  Effective and Efficient Testing of Autonomous Driving Systems},'' in
  \emph{37th IEEE/ACM International Conference on Automated Software
  Engineering}, 2022, pp. 1--8.

\bibitem{ElrofaiScenarioIdentification}
H.~Elrofai, D.~Worm, and O.~Op~den Camp, ``{Scenario Identification for
  Validation of Automated Driving Functions},'' in \emph{Advanced Microsystems
  for Automotive Applications}, 2016, pp. 153--163.

\bibitem{Riedmaier2020}
S.~Riedmaier, T.~Ponn, D.~Ludwig, B.~Schick, and F.~Diermeyer, ``{Survey on
  Scenario-Based Safety Assessment of Automated Vehicles},'' \emph{IEEE
  Access}, vol.~8, pp. 87\,456--87\,477, 2020.

\bibitem{Micha2011}
M.~Lesemann, A.~Zlocki, J.~M. Dalmau, M.~Vesco, H.~Eriksson, J.~Jacobson, and
  L.~Nordstr{\"{o}}m, ``{A test programme for active vehicle safety –
  detailed discussion of the eVALUE testing protocols for longitudinal and
  stability functionality},'' in \emph{22nd Enhanced Safety of Vehicles (ESV)
  Conference}, 2011.

\bibitem{manuel2022}
M.~Mu{\~{n}}oz~S{\'{a}}nchez, J.~Elfring, E.~Silvas, and R.~van~de Molengraft,
  ``{Scenario-based Evaluation of Prediction Models for Automated Vehicles},''
  in \emph{IEEE International Conference on Intelligent Transportation Systems
  (ITSC)}, 2022, pp. 2227--2233.

\bibitem{DeGelder2022}
E.~de~Gelder, J.-P. Paardekooper, A.~Khabbaz~Saberi, H.~Elrofai, O.~Op~den
  Camp, S.~Kraines, J.~Ploeg, and B.~De~Schutter, ``{Towards an Ontology for
  Scenario Definition for the Assessment of Automated Vehicles: An
  Object-Oriented Framework},'' \emph{IEEE Transactions on Intelligent
  Vehicles}, vol.~7, no.~2, pp. 300--314, 2022.

\bibitem{Stellet2015}
J.~E. Stellet, M.~R. Zofka, J.~Schumacher, T.~Schamm, F.~Niewels, and J.~M.
  Zollner, ``{Testing of Advanced Driver Assistance Towards Automated Driving:
  A Survey and Taxonomy on Existing Approaches and Open Questions},'' in
  \emph{IEEE International Conference on Intelligent Transportation Systems
  (ITSC)}, 2015, pp. 1455--1462.

\bibitem{Menzel2018}
T.~Menzel, G.~Bagschik, and M.~Maurer, ``{Scenarios for Development, Test and
  Validation of Automated Vehicles},'' in \emph{IEEE Intelligent Vehicles
  Symposium (IV)}, vol. 2018-June, 2018, pp. 1821--1827.

\bibitem{Schuldt2017}
F.~Schuldt, A.~Reschka, and M.~Maurer, ``{A Method for an Efficient, Systematic
  Test Case Generation for Advanced Driver Assistance Systems in Virtual
  Environments},'' in \emph{Automotive Systems Engineering II}, 2018, pp.
  147--175.

\bibitem{Scholtes2021}
M.~Scholtes, L.~Westhofen, L.~R. Turner, K.~Lotto, M.~Schuldes, H.~Weber,
  N.~Wagener, C.~Neurohr, M.~H. Bollmann, F.~Kortke, J.~Hiller, M.~Hoss,
  J.~Bock, and L.~Eckstein, ``{6-Layer Model for a Structured Description and
  Categorization of Urban Traffic and Environment},'' \emph{IEEE Access},
  vol.~9, pp. 59\,131--59\,147, 2021.

\bibitem{CaiSurvey}
J.~Cai, W.~Deng, H.~Guang, Y.~Wang, J.~Li, and J.~Ding, ``{A Survey on
  Data-Driven Scenario Generation for Automated Vehicle Testing},''
  \emph{Machines}, vol.~10, no.~11, p. 1101, 2022.

\bibitem{WangX2022}
X.~Wang, Y.~Peng, T.~Xu, Q.~Xu, X.~Wu, G.~Xiang, S.~Yi, and H.~Wang,
  ``{Autonomous driving testing scenario generation based on in-depth
  vehicle-to-powered two-wheeler crash data in China},'' \emph{Accident
  Analysis {\&} Prevention}, vol. 176, p. 106812, 10 2022.

\bibitem{Hauer2020}
F.~Hauer, I.~Gerostathopoulos, T.~Schmidt, and A.~Pretschner, ``{Clustering
  Traffic Scenarios Using Mental Models as Little as Possible},'' in \emph{IEEE
  Intelligent Vehicles Symposium (IV)}, 2020, pp. 1007--1012.

\bibitem{CaesarnuScenes}
H.~Caesar, V.~Bankiti, A.~H. Lang, S.~Vora, V.~E. Liong, Q.~Xu, A.~Krishnan,
  Y.~Pan, G.~Baldan, and O.~Beijbom, ``{Nuscenes: A multimodal dataset for
  autonomous driving},'' in \emph{IEEE/CVF Conference on Computer Vision and
  Pattern Recognition (CVPR)}, 2020, pp. 11\,618--11\,628.

\bibitem{Argo}
M.-F. Chang, J.~Lambert, P.~Sangkloy, J.~Singh, S.~Bak, A.~Hartnett, D.~Wang,
  P.~Carr, S.~Lucey, D.~Ramanan, and J.~Hays, ``{Argoverse: 3D tracking and
  forecasting with rich maps},'' in \emph{IEEE/CVF Conference on Computer
  Vision and Pattern Recognition (CVPR)}, 2019, pp. 8740--8749.

\bibitem{Ettinger2021}
S.~Ettinger, S.~Cheng, B.~Caine, C.~Liu, H.~Zhao, S.~Pradhan, Y.~Chai, B.~Sapp,
  C.~Qi, Y.~Zhou, Z.~Yang, A.~Chouard, P.~Sun, J.~Ngiam, V.~Vasudevan,
  A.~McCauley, J.~Shlens, and D.~Anguelov, ``{Large Scale Interactive Motion
  Forecasting for Autonomous Driving: The WAYMO OPEN MOTION DATASET},'' in
  \emph{IEEE/CVF International Conference on Computer Vision (ICCV)}, 2021, pp.
  9690--9699.

\bibitem{MaAppoloScape}
Y.~Ma, X.~Zhu, S.~Zhang, R.~Yang, W.~Wang, and D.~Manocha, ``{TrafficPredict:
  Trajectory prediction for heterogeneous traffic-agents},'' in \emph{AAAI
  Conference on Artificial Intelligence}, vol.~33, no.~01, 2019, pp.
  6120--6127.

\bibitem{Zhao2017ITSC}
D.~Zhao, Y.~Guo, and Y.~J. Jia, ``{TrafficNet: An open naturalistic driving
  scenario library},'' in \emph{IEEE International Conference on Intelligent
  Transportation Systems (ITSC)}, 2017, pp. 1--8.

\bibitem{Waymobehavior}
X.~Hu, Z.~Zheng, D.~Chen, X.~Zhang, and J.~Sun, ``{Processing, assessing, and
  enhancing the Waymo autonomous vehicle open dataset for driving behavior
  research},'' \emph{Transportation Research Part C: Emerging Technologies},
  vol. 134, p. 103490, 1 2022.

\bibitem{GelderRealWorldScenarioMining}
E.~de~Gelder, J.~Manders, C.~Grappiolo, J.~P. Paardekooper, O.~Op~den Camp, and
  B.~De~Schutter, ``{Real-World Scenario Mining for the Assessment of Automated
  Vehicles},'' in \emph{IEEE International Conference on Intelligent
  Transportation Systems (ITSC)}, 2020.

\bibitem{CARLA}
A.~Dosovitskiy, G.~Ros, F.~Codevilla, A.~L{\'{o}}pez, and V.~Koltun, ``{CARLA:
  An Open Urban Driving Simulator},'' in \emph{the 1st Annual Conference on
  Robot Learning}, 2017, pp. 1--16.

\bibitem{Geyer2014}
S.~Geyer, M.~Baltzer, B.~Franz, S.~Hakuli, M.~Kauer, M.~Kienle, S.~Meier,
  T.~Weigerber, K.~Bengler, R.~Bruder, F.~Flemisch, and H.~Winner, ``{Concept
  and development of a unified ontology for generating test and use-case
  catalogues for assisted and automated vehicle guidance},'' \emph{IET
  Intelligent Transport Systems}, vol.~8, no.~3, pp. 183--189, 5 2014.

\bibitem{Bagschik2018}
G.~Bagschik, T.~Menzel, and M.~Maurer, ``{Ontology based Scene Creation for the
  Development of Automated Vehicles},'' \emph{IEEE Intelligent Vehicles
  Symposium (IV)}, pp. 1813--1820, 2018.

\bibitem{iso34501}
{ISO 34501}, ``{Road vehicles — Test scenarios for automated driving systems
  — Vocabulary},'' Internation Organization for Standardization, Tech. Rep.,
  2022.

\bibitem{Erwin2017}
E.~de~Gelder and J.-P. Paardekooper, ``{Assessment of Automated Driving Systems
  using real-life scenarios},'' in \emph{IEEE Intelligent Vehicles Symposium
  (IV)}, 2017, pp. 589--594.

\bibitem{Karunakaran2022HighD}
D.~Karunakaran, J.~S. Berrio, S.~Worrall, and E.~Nebot, ``{Automatic lane
  change scenario extraction and generation of scenarios in OpenX format from
  real-world data},'' \emph{arXiv preprint arXiv:2203.07521}, 2022.

\bibitem{deboor1978practical}
C.~de~Boor, \emph{{A Practical Guide to Splines}}.\hskip 1em plus 0.5em minus
  0.4em\relax Springer, 1 1978, vol.~27.

\bibitem{Pool2017}
E.~A. Pool, J.~F. Kooij, and D.~M. Gavrila, ``{Using road topology to improve
  cyclist path prediction},'' in \emph{IEEE Intelligent Vehicles Symposium
  (IV)}, 2017, pp. 289--296.

\bibitem{Schubert2011}
R.~Schubert, C.~Adam, M.~Obst, N.~Mattern, V.~Leonhardt, and G.~Wanielik,
  ``{Empirical evaluation of vehicular models for ego motion estimation},'' in
  \emph{IEEE Intelligent Vehicles Symposium (IV)}, 2011, pp. 534--539.

\end{thebibliography}
\flushend
\end{document}